%% file: main.tex
\pdfoutput=1
\documentclass[11pt]{article}

\usepackage{acl}

\usepackage{times}
\usepackage{latexsym}

\usepackage[T1]{fontenc}
\usepackage[utf8]{inputenc}

\usepackage{microtype}

\title{Cross-Lingual Knowledge Distillation for\\Answer Sentence Selection in Low-Resource Languages}

\author{Shivanshu Gupta~\thanks{~~This work was done as an intern at Amazon Alexa AI.} \\
  University of California, Irvine \\
  \texttt{shivag5@uci.edu} \\\And
  Yoshitomo Matsubara \\
  Amazon Alexa AI \\
  \texttt{yomtsub@amazon.com} \\\AND
  Ankit Chadha \\
  Amazon Alexa AI \\
  \texttt{ankitrc@amazon.com} \\\And
  Alessandro Moschitti \\
  Amazon Alexa AI \\
  \texttt{amosch@amazon.com} \\
}

\usepackage{graphicx}
\usepackage{adjustbox}
\usepackage{amssymb}
\usepackage{amsmath}
\usepackage{booktabs}
\usepackage{makecell}
\usepackage{enumitem,kantlipsum}
\usepackage{hyperref}
\usepackage{xurl}
\usepackage{algorithm,algpseudocode}
\usepackage{bbm}
\usepackage{caption}
\usepackage{colortbl}
\usepackage{makecell}
\usepackage{multirow}
\usepackage[normalem]{ulem}
\usepackage{xspace}
\usepackage{todonotes}

\newcolumntype{R}[1]{>{\raggedleft\arraybackslash}p{#1}}
\def\localwin{\cellcolor{lightgray}}
\def\highlight{\bf \cellcolor{lightgray}}

\input{commands}

\begin{document}
\maketitle

\begin{abstract}
While impressive performance has been achieved on the task of Answer Sentence Selection (AS2) for English, the same does not hold for languages that lack large labeled datasets. In this work, we propose Cross-Lingual Knowledge Distillation (CLKD) from a strong English AS2 teacher as a method to train AS2 models for low-resource languages in the tasks without the need of labeled data for the target language. To evaluate our method, we introduce 1) \xtrwikiqa,\footnote{\url{https://huggingface.co/datasets/AmazonScience/xtr-wiki_qa}\label{fn:xtrwikiqa-url}} a translation-based WikiQA dataset for 9 additional languages, and 2) \tydiastwo,\footnote{\url{https://huggingface.co/datasets/AmazonScience/tydi-as2}\label{fn:tydias2-url}} a multilingual AS2 dataset with over 70K questions spanning 8 typologically diverse languages. We conduct extensive experiments on \xtrwikiqa and \tydiastwo with multiple teachers, diverse monolingual and multilingual pretrained language models (PLMs) as students, and both monolingual and multilingual training. The results demonstrate that CLKD either outperforms or rivals even supervised fine-tuning with the same amount of labeled data and a combination of machine translation and the teacher model. Our method can potentially enable stronger AS2 models for low-resource languages, while \tydiastwo can serve as the largest multilingual AS2 dataset for further studies in the research community.
\end{abstract}

\input{sections/1-intro.tex}

\input{sections/2-related.tex}

% \input{sections/3-setting.tex}
\input{sections/3-method.tex}

\input{sections/4-datasets.tex}
\input{tables/plms.tex}
\input{tables/wikiqa.tex}
\input{tables/tydias2-translationese.tex}
\input{tables/tydias2-original.tex}
\input{sections/5-experiments.tex}
\input{sections/6-results.tex}

\input{sections/7-conclusion.tex}

\input{sections/8-required.tex}
\bibliography{references-rebiber}
\bibliographystyle{acl_natbib}

\appendix
\input{tables/wikiqa-best_dev_temp}
\input{appendices/dataset-validation.tex}
\input{appendices/best-hyperparam.tex}
\input{tables/tydias2-translationese-best_dev_temp.tex}
\input{tables/tydias2-original-best_dev_temp.tex}
% \section{Pretrained Models}
% \label{sec:pretrained_models}

\end{document}

%% file: commands.tex
\newif\ifcomments
\commentstrue
\ifcomments
    \providecommand{\sg}[1]{{\protect\color{cyan}{[SG: #1]}}}
    \providecommand{\ym}[1]{{\protect\color{orange}{[YM: #1]}}}
    \providecommand{\ac}[1]{{\protect\color{magenta}{[AC: #1]}}}
    \providecommand{\am}[1]{{\protect\color{red}{[AM: #1]}}}

\else
    \providecommand{\sg}[1]{}
    \providecommand{\ym}[1]{}
    \providecommand{\ac}[1]{}
    \providecommand{\am}[1]{}
\fi

\definecolor{applegreen}{rgb}{0.01, 0.65, 0.01}
\definecolor{cardinal}{rgb}{0.77, 0.12, 0.23}

\newcommand{\xtrwikiqa}{Xtr-WikiQA\xspace}
\newcommand{\tydiastwo}{TyDi-AS2\xspace}
\newcommand{\xtrtydiastwo}{Xtr-TyDi-AS2\xspace}
\newcommand{\single}{\textsc{Single}\xspace}
\newcommand{\all}{\textsc{All}\xspace}
\newcommand{\finetune}{\textsc{Finetune}\xspace}
\newcommand{\clkd}{\textsc{Clkd}\xspace}
\newcommand{\clkdr}{\textsc{Clkd[R]}\xspace}
\newcommand{\clkde}{\textsc{Clkd[E]}\xspace}

%% file: sections/1-intro.tex
\section{Introduction}
\label{sec:intro}

\begin{figure}[t]
    \centering
    \includegraphics[width=\linewidth]{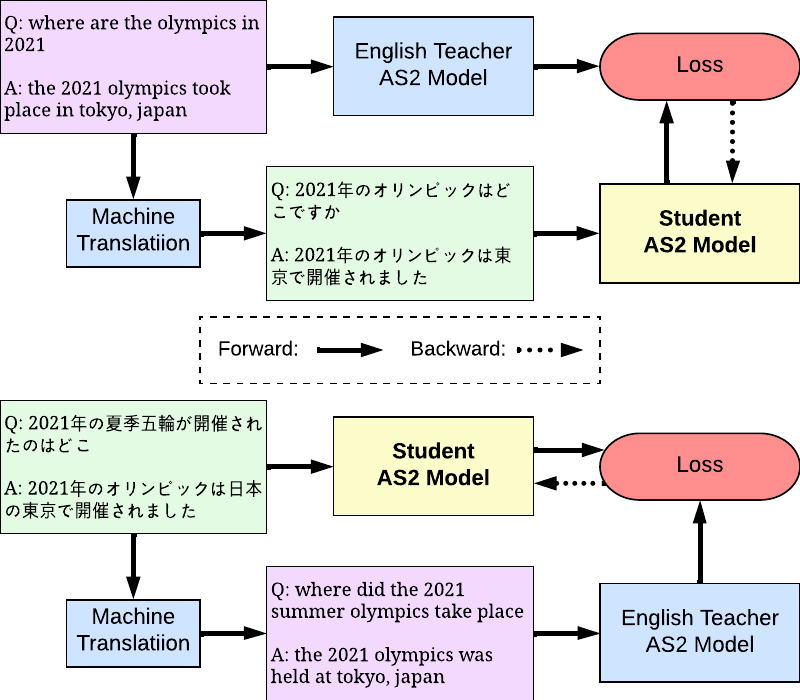}
    \caption{Cross-Lingual Knowledge Distillation (CLKD) in two different scenarios: (\textbf{Top}) using unlabeled English AS2 dataset for target low-resource language lacking any data and (\textbf{Bottom}) using unlabeled original low-resource language AS2 dataset. CLKD enables student AS2 models to learn from English teacher AS2 models without human-annotated datasets.}
    \label{fig:clkd}
\end{figure}

Answer Sentence Selection (AS2) is the task of ranking a given set of answer candidates according to their probability of correctly answering a given question. This is a core task for retrieval-based web Question Answering (QA) systems.
%where the latter play a fundamental roles in many types of real-world applications, e.g., personal assistants, forums, advanced information search systems. 
Indeed, AS2 models applied to the sentences of documents relevant to a question, \emph{e.g.}, retrieved by a search engine, provide accurate answers.

While AS2 has been extensively studied for English~\cite{wang2015long,chen2017benefit,tan2017context,tymoshenko2018cross,nicosia2018semantic,garg2020tanda,tian2020capturing,matsubara2020reranking,laskar2020contextualized,bonadiman2020study,soldaini2020cascade,lauriola2021answer,krishnamurthy2021reference,han2021modeling,zhang2021joint,mrini2021recursive,di2022pre,matsubara2022ensemble}, much less research has been devoted to other languages. This is despite the rapidly increasing importance of multilingual QA with the proliferation of conversational agents and voice assistants using multilingual content from the Web to target locales across the world~\cite{li2022learning}.
A major barrier to achieving similar performance obtained with English in other languages is the lack of large labeled datasets. However, labeling AS2 datasets for every language will be prohibitively expensive as even a single AS2 instance can contain hundreds of candidate answers per question (see Table~\ref{table:tydias2_stats}). This necessitates methods that do not require labeled target language data.
A simple approach is to just translate questions to English and then use an English AS2 model~\cite{vu2021multilingual,asai-etal-2021-xor}. While this pipeline can be quite accurate~\cite{li2022learning}, the need for machine translation makes inference slow and inefficient.
An alternative approach would be to train AS2 models on target language translations of English datasets. However, training using \textit{translationese} seems sub-optimal due to errors and artifacts introduced by machine translation. Moreover, models trained on English questions could be ill-suited for answering target language questions due to information asymmetry~\cite{asai-etal-2021-xor} \emph{i.e.}, questions asked in English are likely to differ from those in the target language due to cultural bias, \emph{e.g.}, they can refer to different entities. 

In this work, we propose Cross-Lingual Knowledge Distillation (CLKD) as a method to use readily available and highly-accurate English AS2 models to train AS2 models for low-resource languages lacking labeled data.
CLKD can use English datasets to train AS2 models for languages lacking any data and can further leverage unlabeled original target language data without the need for costly manual annotation.

Figure~\ref{fig:clkd} illustrates a high-level description of our approach. 
CLKD works similarly to classic Knowledge Distillation (KD)~\cite{hinton2015distilling} in that a student model is trained to mimic a teacher.
The main novelty of our approach is the fact that the teacher and student models operate in different languages, namely the source and target languages. Thus, the input question-answer pairs are translated into both the source and the target languages. Additionally, to allow use of original target language data, which is typically unlabeled, we use only the soft labels obtained from the teacher even when gold labels are available, \emph{i.e.}, given an unlabeled input question and candidate answer pair $(q,a)$, the student is trained using the Kullback-Leibler divergence loss between the probability scores of the teacher and the student when applied to $(q,a)$. 

% i.e., given an unlabeled input question and candidate answer pair $(q,a)$, the student is trained to mimic the teacher using a cross-entropy loss computed between the probability scores of the teacher and the student when applied to $(q,a)$. The main novelty of our approach is the fact that the teacher and student models operate in different languages, namely the source and target language. Thus the input question-answer pairs are translated into both the source (English in our case) and the target language. Additionally, to allow using original target language data, which is typically unlabeled, we do not use loss derived from gold labels but only the soft labels obtained from the teacher.

% CLKD works similarly to standard distillation, i.e., given an unlabeled question/answer pair, $(q,a)$ the student receives a cross-entropy loss computed between the probability scores of the teacher and the student when applied to $(q,a)$. The main novelty of our approach is given by the fact that the teacher is in a different language from the student. Thus, $(q,a)$, which is in the language of the student, is translated into the teacher's language (in our case English), such that the score of the teacher can be computed. Additionally, as we use text from the original language for which there is typically no training AS2 data, we do not use loss derived by labels, i.e., we use the only coming from the teacher.  

To evaluate our approach for diverse languages, we construct two new multilingual AS2 datasets: \xtrwikiqa\footnote{\xtrwikiqa: \textbf{X}-\textbf{tr}anslated\textbf{-WikiQA} \label{fn:xtrwikiqa-name}} (10 languages) and \tydiastwo (8 languages).
\xtrwikiqa consists of 10-language parallel corpora, including the original English corpus from WikiQA~\cite{yang2015wikiqa}.\footnote{\url{https://www.microsoft.com/en-us/download/details.aspx?id=52419}} 
To enable evaluation with original target language data, we further create \tydiastwo by converting TyDi-QA \cite{clark2020tydi},\footnote{\url{https://ai.google.com/research/tydiqa/}} a multilingual QA dataset to an AS2 dataset. We also translate the English \tydiastwo dataset to all the other \tydiastwo languages to build an additional translationese corpus. 
% We make these datasets publicly available

Using the above datasets, we perform extensive experiments with multiple teacher models, about 20 different student models of varying sizes, including both monolingual and multilingual PLMs, and both monolingual and multilingual training. Additionally, to evaluate the utility of our method for both languages lacking any data and for those with some unlabeled data, we experiment with both using only English data (\xtrwikiqa and English \tydiastwo) and only unlabeled target language data (\tydiastwo).
We show that CLKD consistently either rivals or outperforms even supervised fine-tuning with the same amount of gold-labeled data demonstrating the benefit of CLKD using soft labels obtained from a strong English AS2 teacher model.
In particular, we show that CLKD using original language unlabeled data outperforms 1) fine-tuning with gold-labeled translationese and 2) for larger students, even the MT+English AS2 model pipeline, demonstrating the importance of original target language data.

% This shows that our approach can circumvent the errors and artifacts introduced by machine translation and confirms our intuition that models trained on original target language data are better suited for the target language than those trained on translationese.

We expect that the ability of CLKD to train AS2 models without the need for costly annotation process will enable stronger AS2 performance for the world's many low-resource languages. To support further studies on AS2 tasks for such languages, we will make the datasets introduced in this work and our trained models publicly available.

%% file: sections/2-related.tex
\section{Related Work}
We briefly summarize the related studies.

\subsection{KD for Model Compression}
KD was originally proposed as a method for model compression that improves the performance of a weaker model to be trained (student) by learning from a strong but cumbersome model (teacher)~\cite{hinton2015distilling}.
With large pretrained language models based on Transformer~\cite{vaswani2017attention} becoming the new paradigm for natural language processing (NLP) tasks, KD has gained greater attention from the NLP community, with many studies on KD for Transformer-based models~\cite{sanh2019distilbert,jiao2020tinybert,lu2020twinbert,park2021distilling}.
For AS2 tasks,~\citet{matsubara2022ensemble} propose a multi-head student model (CERBERUS) to distill knowledge in an ensemble of multiple diverse teacher models to improve model accuracy without significantly increasing model complexity.

\subsection{Learning from Teacher Models in Different Domains/Tasks}
\citet{garg2021will} propose a technique to filter out non-answerable questions in question-answering systems, which trains a binary classifier for an input question text by mimicking the confidence score from a pretrained AS2 model (input: pair of question and candidate answer).
\citet{gabburo2022knowledge} leverage an AS2 model (a discriminative ranking model) as a teacher model to train an answer generation model~\cite{hsu2021answer}.

There are also a few related studies regarding KD in cross-lingual problem settings.
To address the lack of Chinese sentiment corpora, \citet{wan2009co} leverages machine translations (English-to-Chinese and Chinese-to-English) and studies a cross-lingual sentiment classification problem.
\citet{xu2017cross} also work on sentiment analysis tasks and propose a cross-language distillation with feature adaptation.
\citet{reimers2020making} propose a method to extend existing (English) sentence embedding models to new languages for multilingual student models.
\citet{karamanolakis2020cross} present a text classification model training method with  a small budget of word-level translations for words that are most indicative of the target task and unlabeled documents in the target language.

\citet{li2022learning} propose a multi-stage KD to learn a cross-lingual document retriever from an English retriever, which is the most relevant work to ours.
While similar in regard to learning from an English model, our approach significantly differs from theirs. First, \citet{li2022learning} train cross-lingual document retrievers (\emph{i.e.}, query and document differ in language), whereas we focus on AS2 models taking as input question and answer in the same language, while the teacher is in English and the student is in another language.
Second, while their multi-stage KD method requires the student and teacher to share the embedding size, our single-stage KD method does not have this restriction.
Finally, they evaluate their method on XLM-RoBERTa~\cite{conneau2020unsupervised} only (as student and teacher models), whereas we perform a much more comprehensive study spanning multiple teachers and approximately 20 different pretrained language models.
% \sg{IIRC there were a few other closely related works. We should differentiate from those too, and reduce the focus on \citet{li2022learning}. @Yoshi since I don't have access to my notes from my literature review, could you do this?}
% We find their method most relevant to our approach with respect to learning from an English model.
% However, our approach significantly differs from their method in the following points:
% \begin{enumerate}[topsep=0pt, partopsep=1em, leftmargin=2em, itemsep=-0.1em]
%     \item Their method trains a cross-lingual document retriever (\emph{i.e.}, query and document in different languages) whereas our work uses AS2 models that share a language between input query and candidate answer, and we learn non-English/multilingual models (non-English inputs) from teacher models (English inputs).
%     \item Their multi-stage KD method requires a cross-lingual alignment to support a token-level distillation, which requires student and teacher models to share the embedding size while our singe-stage KD method does not have the limitations.
%     \item While their work assesses the method with XLM-RoBERTa~\cite{conneau2020unsupervised} only, we demonstrate the performance of our method using approximately 20 different pretrained language models (PLMs).
%     % \item We introduce two new datasets to discuss cross-lingual/multilingual question answering systems.
% \end{enumerate}

%% file: sections/3-method.tex
\section{Knowledge Distillation for AS2}
\label{sec:methodology}

\subsection{AS2 Task}
\label{sec:problem_setting}
We consider the task of Answer Sentence Selection where given a question $q$ and a set of answer sentence candidates, $S = \{s_1, \ldots, s_n\}$, the goal is to select the sentence $s^*$ that best answers the question. Following prior work~\cite{garg2020tanda}, we frame this as a ranking task where we assign a score to each sentence $s_i$ for the question $q$ and then select the sentence with the highest score. Formally, given a question-sentence pair $(q, s)$, the AS2 model $\mathcal{M}$ produces a score $\mathcal{M}(q, s)$ measuring the likelihood of $s$ being the correct answer to $q$. We then select the sentence with the highest score as the answer, \emph{i.e.}, $s^* = \operatornamewithlimits{argmax}_{s \in S} \mathcal{M}(q, s)$.
% \sg{removed the Problem Setting section as this is was repeating that.}

\subsection{Knowledge Distillation}
\label{subsec:kd}
Knowledge Distillation~\cite{hinton2015distilling} is an effective method to transfer knowledge from a strong teacher model $T$ to a student model $\mathcal{S}$, by training the student to mimic the teacher. Formally, given inputs $\{x_i\}_{i=1}^N$, the distillation loss is a weighted sum of cross-entropy ($\mathcal{L}_\textnormal{CE}$) of the student w.r.t. gold labels and KL-divergence ($\mathcal{L}_\textnormal{KL}$) of the teacher and student's class probabilities,
% \begin{align} \label{eq:kd_loss}
% \mathcal{L}_{\mathrm{KD}}\left(\theta_S\right) & =-\frac{1}{N} \sum_i^N \mathbf{p}_i^{\left(T\right)} \cdot \log \mathbf{p}_i^{(\mathcal{S})} \\
% \mathbf{p}_i^{\left(T\right)} &= \operatorname{softmax}\left(\mathbf{z}^{\left(T\right)} / \tau\right) \\
% \mathbf{p}_i^{(\mathcal{S})} &= \operatorname{softmax}\left(\mathbf{z}^{(\mathcal{S})} / \tau\right),
% \end{align}
\begin{align}
    \mathcal{L}_\textnormal{KD}(x, y) =& \alpha \mathcal{L}_\textnormal{CE}\left(\operatorname{softmax}\left(\mathbf{z}^{(\mathcal{S})}\right), y\right) + \nonumber\\
    & \hspace{-2em} (1 - \alpha) \tau^2 \mathcal{L}_\textnormal{KL}(\mathbf{p}_i^{\left(T\right)}, \mathbf{p}_i^{(\mathcal{S})})
    \label{eq:kd_loss}, \\ \nonumber
    \mathbf{p}_i^{\left(T\right)} =& \operatorname{softmax}\left(\mathbf{z}^{\left(T\right)} / \tau\right) \nonumber \\
    \mathbf{p}_i^{(\mathcal{S})} =& \operatorname{softmax}\left(\mathbf{z}^{(\mathcal{S})} / \tau\right), \nonumber
\end{align}
\noindent where $\mathbf{z}_i^{\left(T\right)} = T(x_i)$ and $\mathbf{z}_i^{\left(\mathcal{S}\right)}=\mathcal{S}(x_i)$ are logits from teacher and student models, respectively.
$y$ indicates a gold label (human annotation), and $\alpha$ and $\tau$ (``temperature'') are hyperparameters.
% We describe the hyperparameter tuning in \S~\ref{subsec:training}.

\subsection{Cross-Lingual Knowledge Distillation}
\label{subsec:clkd}
In order to train AS2 models for low-resource languages lacking labeled data, we propose Cross-Lingual Knowledge Distillation (CLKD) from accurate and readily available English AS2 models. CLKD assumes the absence of gold labels for target languages and, in general, teaches the student model for a ``target'' language to mimic the teacher from a different ``source'' language (English in this study) as illustrated in Fig.~\ref{fig:clkd}. In other words, the CLKD loss is the second term of Eq.~\ref{eq:kd_loss} ($\alpha = 0$, no gold labels are used) with student and teacher logits obtained by feeding them the same input in target and source language, respectively.
% Unlike the classic monolingual knowledge distillation described above, CLKD is supposed to solve the scarcity or even complete absence of data in the target language, the student model for the ``target'' language(s) needs to mimic a teacher from a different ``source'' language when fed parallel inputs.
% In other words, we set $\alpha = 0$ in Eq. (\ref{eq:kd_loss}) for CLKD, since human annotations $y$ are typically not available in our scenario.
% \todo{I need to revise this part below as it is not enough clear. The concept is super simple and can be explained very quickly, here the notation seems a bit involuted. I will give a better read and see if it can be simplified}

Formally, given a teacher $T^l$ for source language $l$, and two parallel unlabeled datasets, $D^{(l)} = \left\{x_i^{(l)}\right\}_{i=1}^{N}$ and $D^{(l')} = \left\{x_i^{(l')}\right\}_{i=1}^{N}$, for source and target languages, $l$ and $l'$, respectively, CLKD trains a student model $\mathcal{S}^{l'}$ using the same loss as the monolingual distillation case (Eq.~\ref{eq:kd_loss}) but with the teacher and student logits obtained as $\mathbf{z}_i^{\left(T\right)} = T(x_i^{(l)})$ and $\mathbf{z}_i^{\left(\mathcal{S}\right)}=\mathcal{S}(x_i^{(l')})$, respectively.

For the AS2 task, the input will be question-sentence pairs \emph{i.e.}, $x_i = (q_i, s_i)$. Additionally, as we distill knowledge in English AS2 models, the source and target languages will be English and a low-resource language, respectively. Also, since parallel datasets are likely not available, they will be obtained using automatic machine translation.

Depending on the low-resource language data available, CLKD can be applied in two different ways:
(1) In absence of any target-language data, CLKD can be applied using an English AS2 dataset (see Fig.~\ref{fig:clkd} top). In this scenario, the teacher and student will be fed the original English and translationese instances, respectively. While this can be applied to any language, errors and artifacts inevitably introduced by machine translation and information asymmetry due to cultural bias with respect to the target language \cite{asai-etal-2021-xor} will limit the student's performance. 
(2) CLKD allows overcoming this limitation by utilising original target language unlabeled data when available. As shown in the Fig.~\ref{fig:clkd} bottom, this would involve feeding the original language and English-translated input to the student and teacher models, respectively.

Note that the success of CLKD, particularly with original data, relies on two practical assumptions: (i) two AS2 models for two different languages should produce similar probability scores when applied to inputs that are translations of each other, and (ii) the teacher working on automatically translated data is still accurate enough to transfer \emph{useful} knowledge to the student.
% As we will see \S~\ref{sec:evaluations}, this is indeed the case.

% Denoting translation from $l$ to $l'$ as $l \rightarrow l'$, we use $D^{(l \rightarrow l')} = \left\{\left(q_i^{(l \rightarrow l')}, s_i^{(l \rightarrow l')}\right)\right\}_{i=1}^{N}$ to denote the parallel AS2 dataset obtained by translating the language $l$ dataset $D^{(l)} = \left\{\left(q_i^{(l)}, s_i^{(l)}\right)\right\}_{i=1}^{N}$ to $l'$.
% Finally, for low-resource languages for which an unlabeled dataset exists, CLKD will be applied by translating it to English \emph{i.e.}, using $D^{(l' \rightarrow en)}$ and $D^{(l')}$.
% Conversely, we can also translate English AS2 datasets to the target language.
% This will yield a student model trained on translationese but can be applied to languages lacking even unlabeled data. \todo{invert the order of describing the two settings.}

%% file: sections/4-datasets.tex
\input{tables/wikiqa-statistics.tex}

\input{tables/tydias2-statistics.tex}

\section{New Datasets}
In this section, we introduce two new AS2 datasets, \xtrwikiqa and \tydiastwo.
The datasets are constructed from the WikiQA \cite{yang2015wikiqa} and TyDi-QA \cite{clark2020tydi} datasets, respectively, and the intended use of our new datasets follows Community Data License Agreement (CDLA) - Permissive (Version 2.0).\footnote{\url{https://cdla.dev/permissive-2-0/}}

\subsection{\xtrwikiqa}
\label{subsec:xtrwikiqa}
WikiQA~\cite{yang2015wikiqa} has been used as an English AS2 dataset in various studies on AS2 tasks~\cite{garg2020tanda,matsubara2020reranking,lauriola2021answer}.
Following~\cite{garg2020tanda,matsubara2022ensemble}, we remove queries which have no correct answers from the training split, but leave such queries in the development and test splits.

We translate WikiQA using Amazon Translate\footnote{\url{https://aws.amazon.com/translate/} \label{fn:amazon_translate}} to construct a new multilingual AS2 dataset, named \xtrwikiqa,\textsuperscript{\ref{fn:xtrwikiqa-name}} comprising 9 additional languages (\emph{i.e.}, 10 languages in total): Arabic (ar), Dutch (nl), French (fr), German (de), Hindi (hi), Italian (it), Japanese (ja), Portuguese (pt), and Spanish (es).
Table~\ref{table:wikiqa_stats} shows the statistics of \xtrwikiqa dataset.
Each of the 10 language corpora in \xtrwikiqa has the same statistics as those are parallel corpora.

\subsection{\tydiastwo}
\label{subsec:tydias2}
% Although the quality of automatic translation has reached very high accuracy, the evaluation of our approach should be carried out on natural language, not on translationese.
In addition to our \emph{translationese dataset} above, we need a large and accurate multilingual AS2 dataset to evaluate our method and compare against supervised baselines on original target language data.
Due to the lack of such datasets, we introduce \tydiastwo, a large multilingual AS2 benchmark derived from the TyDi-QA dataset~\cite{clark2020tydi}, a multilingual Machine Reading dataset.
\tydiastwo is a collection of AS2 datasets for eight typologically diverse languages, including Bengali (bn), English (en), Finnish (fi), Indonesian (id), Japanese (ja), Korean (ko), Russian (ru), and Swahili (sw).
The dataset was constructed from the data for the primary task in TyDi-QA, where each instance is accompanied by a Wikipedia article.

\paragraph{Conversion} 
TyDi-QA is a QA dataset spanning questions from 11 typologically diverse languages.
Each instance comprises a human-generated question, a single Wikipedia document as context, and one or more spans from the document containing the answer.
To convert each instance into AS2 instances, we split the context document into sentences and use the answer spans to identify the correct answer sentences. 
% (See Appendix~\ref{sec:dataset-validation}.)

To split documents, we use multiple different sentence tokenizers for the diverse languages and omit languages for which we could not find a suitable sentence tokenizer: 1) \texttt{bltk}\footnote{\url{https://github.com/saimoncse19/bltk}} for Bengali, 2) \texttt{blingfire}\footnote{\url{https://github.com/microsoft/BlingFire}} for Swahili, Indonesian, and Korean, 3) \texttt{pysdb}\footnote{\url{https://github.com/nipunsadvilkar/pySBD}}~\cite{sadvilkar2020pysbd} for English and Russian, 4) \texttt{nltk}\footnote{\url{https://www.nltk.org/}}~\cite{bird2009natural} for Finnish, and 5) \texttt{Konoha}\footnote{\url{https://github.com/himkt/konoha}} for Japanese.

\paragraph{Translation} 
For CLKD experiments with original target language data, we use Amazon Translate\textsuperscript{\ref{fn:amazon_translate}} to translate the non-English corpora of \tydiastwo datasets into English.
Furthermore, to conduct another translationese experiments, we also translate the English \tydiastwo dataset to all the other \tydiastwo languages, similar to~\cite{vu2021multilingual}. We refer to this dataset as \xtrtydiastwo.

% To additionally discuss the effect of our proposed approach on translationese data, we also create the translationese version of \tydiastwo by translating its English corpus with Amazon Translate into non-English, which is similar to~\cite{vu2021multilingual}. We'll refer to this  

% \paragraph{Validation} As both sentence tokenization and answer sentence identification are not perfect, we manually validated the newly created AS2 datasets. We randomly sampled 50 instances for each language and manually checked the correctness of the answer sentences and found the annotated correct answer sentence to be correct in 98\% of the cases. \sg{@yoshi any reason this was removed?}
% \sg{TODO: add validation statistics}

\paragraph{Dataset Statistics} As the original TyDi-QA test set is not publicly available, we repurposed the dev set for test set and used an 80-20 split of the original training set to create \tydiastwo's training and dev sets.
Table \ref{table:tydias2_stats} shows statistics of \tydiastwo.

%% file: tables/wikiqa-statistics.tex
\begin{table}[t]
    \centering
    \bgroup
    % \small
    % \setlength{\tabcolsep}{0.3em}
% 	\renewcommand{\arraystretch}{1.15}
    \begin{tabular}{@{}lrrr@{}}
        \toprule
        & \multicolumn{1}{c}{\textbf{train}} & \multicolumn{1}{c}{\textbf{dev}} & \multicolumn{1}{c}{\textbf{test}} \\
        \midrule
        \textbf{\#Queries} & 873 & 126 & 243 \\
        \textbf{\#QA pairs} & 8,671 & 1,130 & 2,351 \\
        \textbf{\#Correct answers} & 1,040 & 140 & 293 \\
        \bottomrule
    \end{tabular}
    \egroup
    \caption{Statistics of \xtrwikiqa for each language.}
    \label{table:wikiqa_stats}
\end{table}

%% file: tables/tydias2-statistics.tex
\begin{table*}[t]
    \centering
    \bgroup
    \small
    \setlength{\tabcolsep}{0.3em}
    \begin{tabular}{@{}lR{3.3em}R{3.3em}R{3.3em}R{4em}R{3.7em}R{3.4em}R{2.5em}R{2.5em}R{2.5em}R{2.5em}R{2.5em}R{2.5em}@{}}
        \toprule
        \multicolumn{1}{c}{\multirow{2}{*}{\textbf{Language}}} & \multicolumn{3}{c}{\textbf{\#Queries}} & \multicolumn{3}{c}{\textbf{\#Sentences}} & \multicolumn{3}{l}{\textbf{Avg. Sentence Length}} & \multicolumn{3}{c}{\textbf{\#Positive QA Pairs}} \\
        % \cline{2-13}
        & \multicolumn{1}{c}{\textbf{train}} & \multicolumn{1}{c}{\textbf{dev}} & \multicolumn{1}{c}{\textbf{test}} & \multicolumn{1}{c}{\textbf{train}} & \multicolumn{1}{c}{\textbf{dev}} & \multicolumn{1}{c}{\textbf{test}} & \multicolumn{1}{c}{\textbf{train}} & \multicolumn{1}{c}{\textbf{dev}} & \multicolumn{1}{c}{\textbf{test}} & \multicolumn{1}{c}{\textbf{train}} & \multicolumn{1}{c}{\textbf{dev}} & \multicolumn{1}{c}{\textbf{test}} \\
        \midrule
        \textbf{Bengali (bn)} & 7,978 & 2,056 & 316 &  1,376,432 & 351,186 & 37,465 & 106.3 & 106.3 & 106.7 & 1,914 & 472 & 148 \\
        \textbf{English (en)} & 6,730 & 1,686 & 918 & 1,643,702 & 420,899 & 249,513 & 107.8 & 106.6 & 107.4 & 2,953 & 699 & 810 \\
        \textbf{Finnish (fi)} & 10,859 & 2,731 & 1,870 & 1,567,695 & 408,205 & 298,093 & 123.3 & 122.4 & 123.5 & 5,317 & 1,316 & 1,211 \\
        \textbf{Indonesian (id)} & 9,310 & 2,339 & 1,355 & 960,270 & 236,076 & 97,057 & 154.6 & 155.3 & 153.9 & 2,237 & 608 & 408 \\
        \textbf{Japanese (ja)} & 11,848 & 2,981 & 1,504 & 3,183,037 & 822,654 & 444,106 & 45.2 & 45.2 & 46.1 & 3,513 & 846 & 858 \\
        \textbf{Korean (ko)} & 7,354 & 1,943 & 1,389 & 1,558,191 & 392,361 & 199,043 & 84.2 & 84.2 & 88.4 & 586 & 141 & 216 \\
        \textbf{Russian (ru)} & 9,187 & 2,294 & 1,395 & 3,190,650 & 820,668 & 367,595 & 109.3 & 110.0 & 101.6 & 5,101 & 1,277 & 1,039 \\
        \textbf{Swahili (sw)} & 8,350 & 2,850 & 1,896 & 1,048,303 & 269,894 & 74,775 & 145.3 & 144.0 & 141.0 & 976 & 244 & 356 \\
        \bottomrule
    \end{tabular}
    \egroup
    \caption{Statistics of \tydiastwo.}
    \label{table:tydias2_stats}
\end{table*}

%% file: tables/plms.tex
\begin{table*}[t]
    \centering
    \small
    \bgroup
    \setlength{\tabcolsep}{0.2em}
    \renewcommand{\arraystretch}{1.15}
    \begin{tabular}{@{}cclll@{}}
        \toprule
        \multicolumn{1}{c}{\textbf{Language}} & \multicolumn{1}{c}{\textbf{Dataset}} & \multicolumn{1}{c}{\textbf{Hugging Face Pretrained Model}} & \multicolumn{1}{c}{\textbf{Size}} & \multicolumn{1}{c}{\textbf{Note}} \\
        \midrule
        en & \xtrwikiqa & \href{https://huggingface.co/roberta-large}{roberta-large} & 355M & \makecell{\cite{liu2019roberta}\\Fine-tuned by~\citet{garg2020tanda}} \\
        en & \makecell{\xtrwikiqa\\\tydiastwo} & \href{https://huggingface.co/google/electra-large-discriminator}{google/electra-large-discriminator} & 335M & \makecell{\cite{clark2019electra}\\Fine-tuned by~\citet{matsubara2022ensemble}} \\
        \midrule
        ar & \xtrwikiqa & \href{https://huggingface.co/asafaya/bert-base-arabic}{asafaya/bert-base-arabic} & 111M & \cite{safaya2020kuisail} \\
        de & \xtrwikiqa & \href{https://huggingface.co/bert-base-german-cased}{bert-base-german-cased} & 109M & \\
        hi & \xtrwikiqa & \href{https://huggingface.co/monsoon-nlp/hindi-bert}{monsoon-nlp/hindi-bert} & 14.7M & \\
        it & \xtrwikiqa & \href{https://huggingface.co/dbmdz/bert-base-italian-xxl-cased}{dbmdz/bert-base-italian-xxl-cased} & 111M & \\
        ja & \makecell{\xtrwikiqa\\\tydiastwo} & \href{https://huggingface.co/nlp-waseda/roberta-base-japanese}{nlp-waseda/roberta-base-japanese} & 111M & \\
        nl & \xtrwikiqa & \href{https://huggingface.co/GroNLP/bert-base-dutch-cased}{GroNLP/bert-base-dutch-cased} & 109M & \cite{de2019bertje} \\
        pt & \xtrwikiqa & \href{https://huggingface.co/neuralmind/bert-base-portuguese-cased}{neuralmind/bert-base-portuguese-cased} & 109M & \cite{souza2020bertimbau}\\
        
        bn & \tydiastwo & \href{https://huggingface.co/csebuetnlp/banglabert}{csebuetnlp/banglabert} & 111M & \cite{bhattacharjee2022banglabert} \\
        fi & \tydiastwo & \href{https://huggingface.co/TurkuNLP/bert-base-finnish-cased-v1}{TurkuNLP/bert-base-finnish-cased-v1} & 125M & \cite{virtanen2019multilingual} \\
        id & \tydiastwo & \href{https://huggingface.co/indobenchmark/indobert-base-p1}{indobenchmark/indobert-base-p1} & 124M & \cite{wilie2020indonlu} \\
        % ja & \tydiastwo & \href{https://huggingface.co/nlp-waseda/roberta-base-japanese}{nlp-waseda/roberta-base-japanese} & 111M & \\
        ko & \tydiastwo & \href{https://huggingface.co/klue/bert-base}{klue/bert-base} & 111M & \cite{park2021klue} \\
        ru & \tydiastwo & \href{https://huggingface.co/DeepPavlov/rubert-base-cased}{DeepPavlov/rubert-base-cased} & 178M & \cite{kuratov2019adaptation} \\
        sw & \tydiastwo & \href{https://huggingface.co/Davlan/bert-base-multilingual-cased-finetuned-swahili}{Davlan/bert-base-multilingual-cased-finetuned-swahili} & 178M & Fine-tuned on Swahili corpus \\
        \midrule
        multi & \makecell{\xtrwikiqa\\\tydiastwo} & \href{https://huggingface.co/bert-base-multilingual-cased}{bert-base-multilingual-cased} & 178M & \cite{devlin2019bert} \\
        multi & \makecell{\xtrwikiqa\\\tydiastwo} & \href{https://huggingface.co/xlm-roberta-base}{xlm-roberta-base} & 278M & \cite{conneau2020unsupervised} \\
        multi & \makecell{\xtrwikiqa\\\tydiastwo} & \href{https://huggingface.co/xlm-roberta-large}{xlm-roberta-large} & 560M & \cite{conneau2020unsupervised} \\
        \bottomrule
    \end{tabular}
    \egroup
    \caption{List of pretrained language models used in this study.}
    \label{table:pretrained_models}
\end{table*}

%% file: tables/wikiqa.tex
\begin{table*}[t]
    \centering
    \bgroup
    \setlength{\tabcolsep}{0.45em}
    % \small
    \begin{tabular}{lllrrrrrrrrr}
        \toprule
        \multicolumn{1}{c}{\textbf{Language}} & \multicolumn{1}{c}{\textbf{Student LM}} & \multicolumn{1}{c}{\textbf{Method}} & \multicolumn{1}{c}{\textbf{ar}} & \multicolumn{1}{c}{\textbf{de}} & \multicolumn{1}{c}{\textbf{es}} & \multicolumn{1}{c}{\textbf{fr}} & \multicolumn{1}{c}{\textbf{hi}} & \multicolumn{1}{c}{\textbf{it}} & \multicolumn{1}{c}{\textbf{ja}} & \multicolumn{1}{c}{\textbf{nl}} & \multicolumn{1}{c}{\textbf{pt}} \\
        \midrule
        % English & \multicolumn{2}{l}{TEACHER: ELECTRA-Large} & 87.7 & 87.7 & 87.7 & 87.7 & 87.7 & 87.7 & 87.7 & 87.7 & 87.7 \\
        % ~ & \multicolumn{2}{l}{TEACHER: RoBERTa-Large} & 91.8 & 91.8 & 91.8 & 91.8 & 91.8 & 91.8 & 91.8 & 91.8 & 91.8 \\
        % \midrule
        \single & Monolingual & \finetune & 56.2 & 63.4 & N/A & N/A & 27.2 & 63.8 & 57.8 & 59.8 & 63.7 \\ 
        ~ & ~ & \clkde & \localwin 63.9 & \localwin 70.0 & N/A & N/A & \localwin 27.3 & \localwin 66.0 & \localwin 58.0 & \localwin 65.4 & \localwin 68.6 \\ 
        ~ & ~ & \clkdr & \localwin 65.7 & \localwin 72.0 & N/A & N/A & \localwin 28.4 & \localwin 70.1 & 57.6 & \localwin 66.7 & \localwin 68.6 \\
        \cline{2-12}
        ~ & mBERT & \finetune & 61.9 & 66.0 & 68.0 & 63.8 & 61.3 & 60.8 & 66.1 & 63.5 & 61.7 \\ 
        ~ & ~ & \clkde & \localwin 69.1 & \localwin 72.6 & \localwin 75.0 & \localwin 71.2 & \localwin 65.6 & \localwin 73.4 & \localwin 71.6 & \localwin 71.5 & \localwin 74.9 \\ 
        ~ & ~ & \clkdr & \localwin 69.3 & \localwin 73.1 & \localwin 75.2 & \localwin 71.5 & \localwin 68.3 & \localwin 75.7 & \localwin 71.7 & \localwin 74.9 & \localwin 74.2 \\ 
        \cline{2-12}
        ~ & XLM-R-Base & \finetune & 56.5 & 56.5 & 59.9 & 57.9 & 56.2 & 58.2 & 59.0 & 31.3 & 56.8 \\ 
        ~ & ~ & \clkde & \localwin 66.3 & \localwin 64.6 & \localwin 71.5 & \localwin 65.2 & \localwin 64.1 & \localwin 68.2 & \localwin 66.0 & \localwin 65.8 & \localwin 70.0 \\ 
        ~ & ~ & \clkdr & \localwin 67.5 & \localwin 64.3 & \localwin 69.1 & \localwin 66.8 & \localwin 65.8 & \localwin 68.5 & \localwin 68.9 & \localwin 58.7 & \localwin 70.0 \\ 
        \cline{2-12}
        ~ & XLM-R-Large & \finetune & 64.2 & 74.1 & 71.1 & 69.1 & 71.6 & 66.9 & 71.3 & 71.1 & 75.3 \\ 
        ~ & ~ & \clkde & \localwin 76.3 & \localwin 81.5 & \localwin 81.3 & \localwin 80.0 & \localwin 79.6 & \localwin 80.9 & \localwin 79.6 & \localwin 81.5 & \localwin 82.7 \\ 
        ~ & ~ & \clkdr & \localwin 76.3 & \localwin 81.9 & \localwin 81.6 & \localwin 81.9 & \localwin 80.3 & \localwin 80.7 & \localwin 80.4 & \localwin 80.7 & \localwin 81.9 \\
        \hline
        \all & mBERT & \finetune & 70.8 & 70.2 & 74.1 & 69.3 & 67.8 & 71.6 & 69.3 & 70.4 & 74.2 \\ 
        ~ & ~ & \clkde & \localwin 74.6 & \localwin 77.1 & \localwin 79.3 & \localwin 76.3 & \localwin 71.1 & \localwin 78.7 & \localwin 74.9 & \localwin 75.5 & \localwin 80.0 \\ 
        ~ & ~ & \clkdr & \localwin 76.3 & \localwin 77.9 & \localwin 80.8 & \localwin 75.5 & \localwin 71.3 & \localwin 80.0 & \localwin 76.7 & \localwin 77.6 & \localwin 80.4 \\ 
        \cline{2-12}
        ~ & XLM-R-Base & \finetune & 59.1 & 63.0 & 66.5 & 63.9 & 61.9 & 64.6 & 61.6 & 63.1 & 64.3 \\ 
        ~ & ~ & \clkde & \localwin 71.6 & \localwin 72.6 & \localwin 73.4 & \localwin 70.9 & \localwin 70.7 & \localwin 74.5 & \localwin 68.7 & \localwin 69.8 & \localwin 75.0 \\ 
        ~ & ~ & \clkdr & \localwin 73.8 & \localwin 73.9 & \localwin 74.1 & \localwin 70.1 & \localwin 70.2 & \localwin 73.5 & \localwin 69.3 & \localwin 71.3 & \localwin 75.0 \\ 
        \cline{2-12}
        ~ & XLM-R-Large & \finetune & 75.6 & 82.9 & 78.2 & 80.0 & 78.9 & 79.4 & 78.5 & 80.4 & 81.8 \\ 
        ~ & ~ & \clkde & \localwin 80.3 & \localwin 82.7 & \localwin 81.9 & \localwin 82.5 & \localwin 81.3 & \highlight 83.8 & \highlight 82.5 & \highlight 83.5 & \localwin 85.0 \\ 
        ~ & ~ & \clkdr & \highlight 81.6 & \highlight 82.6 & \highlight 84.2 & \highlight 84.5 & \highlight 80.4 & \localwin 82.9 & \localwin 81.8 & \localwin 82.0 & \highlight 85.1 \\
        \bottomrule
    \end{tabular}
    \egroup
    \caption{\textbf{\xtrwikiqa}: Averaged test results (P@1) of models trained on dataset of 1) single target language and 2) all the target languages. We highlight \adjustbox{bgcolor=lightgray}{\strut better results(\finetune vs. \clkd)} and additionally use a bold font for \adjustbox{bgcolor=lightgray}{\strut\bf the best student model} for each of the nine target languages. Our English teacher models, ELECTRA-Large (E) and RoBERTa-Large (R), achieved 87.7\% and 91.8\% P@1 respectively.}
    \label{table:wikiqa}
\end{table*}

%% file: tables/tydias2-translationese.tex
\begin{table*}[t]
    \centering
    \begin{tabular}{lllrrrrrrr}
        \toprule
        \multicolumn{1}{c}{\textbf{Language}} & \multicolumn{1}{c}{\textbf{Student LM}} & \multicolumn{1}{c}{\textbf{Method}} & \multicolumn{1}{c}{\textbf{bn}} & \multicolumn{1}{c}{\textbf{fi}} & \multicolumn{1}{c}{\textbf{id}} & \multicolumn{1}{c}{\textbf{ja}} & \multicolumn{1}{c}{\textbf{ko}} & \multicolumn{1}{c}{\textbf{ru}} & \multicolumn{1}{c}{\textbf{sw}} \\
        \midrule
        \multicolumn{3}{c}{\textsc{MT + Teacher}} & 63.9 & 69.2 & 81.0 & 55.4 & 77.8 & 66.8 & 86.4 \\
        \hline
        \single & Monolingual & \finetune & 34.6 & 54.6 & 66.1 & 24.6 & 71.6 & 53.7 & 66.9 \\ 
        ~ & ~ & \clkde & \localwin 35.6 & \localwin 59.7 & \localwin 70.6 & \localwin 26.8 & \localwin 74.7 & \localwin 59.0 & \localwin 69.7 \\ 
        \cline{2-10}
        ~ & mBERT & \finetune & 37.4 & 52.4 & 69.7 & 31.6 & 70.9 & 51.7 & 72.7 \\ 
        ~ & ~ & \clkde & \localwin 43.3 & \localwin 58.6 & \localwin 74.1 & \localwin 35.4 & \localwin 79.5 & \localwin 55.6 & \localwin 74.8 \\ 
        \cline{2-10}
        ~ & XLM-R-Base & \finetune & 29.0 & 48.8 & 69.7 & 33.0 & 64.9 & 47.3 & 67.0 \\ 
        ~ & ~ & \clkde & \localwin 34.4 & \localwin 53.9 & \localwin 72.1 & \localwin 37.8 & \localwin 69.3 & \localwin 51.3 & \localwin 70.1 \\ 
        \cline{2-10}
        ~ & XLM-R-Large & \finetune & 54.9 & 61.3 & 77.3 & 52.0 & 75.7 & 59.2 & 85.3 \\ 
        ~ & ~ & \clkde & \localwin 61.3 & \localwin 65.0 & \localwin 80.9 & \localwin 55.8 & \localwin 78.2 & \localwin 62.2 & \localwin 85.0 \\
        \hline
        \all & mBERT & \finetune & 51.0 & 57.0 & 74.8 & 46.3 & 73.9 & 56.3 & 76.9 \\ 
        ~ & ~ & \clkde & \localwin 54.9 & \localwin 61.8 & \localwin 77.4 & \localwin 49.6 & \localwin 80.3 & \localwin 60.3 & \localwin 78.8 \\ 
        \cline{2-10}
        ~ & XLM-R-Base & \finetune & 43.1 & 53.9 & 72.6 & 40.4 & 70.4 & 52.5 & 73.0 \\ 
        ~ & ~ & \clkde & \localwin 50.5 & \localwin 57.9 & \localwin 75.6 & \localwin 44.8 & \localwin 77.2 & \localwin 56.4 & \localwin 77.1 \\ 
        \cline{2-10}
        ~ & XLM-R-Large & \finetune & 62.8 & 63.9 & 80.7 & 57.5 & 77.3 & 62.3 & 83.9 \\ 
        ~ & ~ & \clkde & \highlight 67.2 & \highlight 67.5 & \highlight 81.8 & \highlight 58.4 & \highlight 81.1 & \highlight 66.7 & \highlight 85.5 \\
        \bottomrule
    \end{tabular}
    \caption{\textbf{\xtrtydiastwo}: Averaged test results (P@1) of models trained on \uline{translationese} data for 1) single target language and 2) all the target languages. We highlight \adjustbox{bgcolor=lightgray}{\strut better results (\finetune vs. \clkd)} and additionally use a bold font for \adjustbox{bgcolor=lightgray}{\strut\bf the best student model} for each of the seven target languages.}
    \label{table:tydias2-translationese}
\end{table*}

%% file: tables/tydias2-original.tex
\begin{table*}[t]
    \centering
    \begin{tabular}{lllrrrrrrrr}
        \toprule
        \multicolumn{1}{c}{\textbf{Language}} & \multicolumn{1}{c}{\textbf{Student LM}} & \multicolumn{1}{c}{\textbf{Method}} & \multicolumn{1}{c}{\textbf{bn}} & \multicolumn{1}{c}{\textbf{fi}} & \multicolumn{1}{c}{\textbf{id}} & \multicolumn{1}{c}{\textbf{ja}} & \multicolumn{1}{c}{\textbf{ko}} & \multicolumn{1}{c}{\textbf{ru}} & \multicolumn{1}{c}{\textbf{sw}} \\
        \midrule
        \multicolumn{3}{c}{\textsc{MT + Teacher}} & 63.9 & 69.2 & 81.0 & 55.4 & 77.8 & 66.8 & 86.4 \\
        \hline
        \single & Monolingual & \finetune & \localwin 54.9 & 61.8 & 71.2 & 30.1 & 71.6 & \localwin 65.0 & 71.3 \\ 
        ~ & ~ & \clkde & 53.1 & \localwin 63.0 & \localwin 72.8 & \localwin 34.0 & \localwin 77.7 & 63.6 & \localwin 74.5 \\ 
        \cline{2-10}
        ~ & mBERT & \finetune & 50.5 & 60.3 & 75.5 & 53.2 & 70.9 & \localwin 60.1 & 73.8 \\ 
        ~ & ~ & \clkde & \localwin 53.6 & \localwin 61.6 & \localwin 78.6 & \localwin 57.0 & \localwin 75.9 & 58.5 & \localwin 78.1 \\ 
        \cline{2-10}
        ~ & XLM-R-Base & \finetune & 50.3 & 53.8 & 71.0 & 45.4 & 69.5 & \localwin 54.9 & 69.9 \\ 
        ~ & ~ & \clkde & \localwin 51.5 & \localwin 56.8 & \localwin 77.6 & \localwin 49.9 & \localwin 74.2 & 54.0 & \localwin 77.5 \\ 
        \cline{2-10}
        ~ & XLM-R-Large & \finetune & 66.4 & \localwin 68.7 & 78.3 & \localwin 59.3 & 75.4 & 65.6 & 84.3 \\ 
        ~ & ~ & \clkde & \localwin 67.7 & 67.0 & \localwin 81.9 & 57.7 & \localwin 80.3 & \localwin 66.4 & \localwin 86.1 \\ 
        \hline
        \all & mBERT & \finetune & 58.5 & \localwin 65.1 & 80.4 & 56.6 & 76.7 & \localwin 63.4 & 80.6 \\ 
        ~ & ~ & \clkde & \localwin 59.2 & 63.4 & \localwin 81.8 & \localwin 57.1 & \localwin 81.9 & 62.6 & \localwin 82.3 \\ 
        \cline{2-10}
        ~ & XLM-R-Base & \finetune & 55.4 & \localwin 60.7 & 77.1 & 50.1 & 73.2 & 59.2 & 80.4 \\ 
        ~ & ~ & \clkde & \localwin 56.9 & 59.8 & \localwin 78.0 & \localwin 52.7 & \localwin 78.0 & \localwin 59.7 & \localwin 83.1 \\ 
        \cline{2-10}
        ~ & XLM-R-Large & \finetune & \highlight 70.0 & \highlight 72.3 & 82.2 & \highlight 62.9 & 80.5 & \highlight 68.4 & \highlight 88.6 \\ 
        ~ & ~ & \clkde & 68.0 & 68.8 & \highlight 84.0 & 58.0 & \highlight 83.4 & 68.3 & 87.1 \\
        \bottomrule
    \end{tabular}
    \caption{\textbf{\tydiastwo}: Averaged test results (P@1) of models trained on \uline{original language} data for 1) single target language and 2) all the target languages. We highlight \adjustbox{bgcolor=lightgray}{\strut better results (\finetune vs. \clkd)} and additionally use a bold font for \adjustbox{bgcolor=lightgray}{\strut\bf the best student model} for each of the seven target languages.}
    \label{table:tydias2-original}
\end{table*}

%% file: sections/5-experiments.tex
\section{Experimental Setup}
\label{sec:experimental_setup}

For a rigorous assessment of the efficacy of CLKD, we design various experiments with different teachers, students, and training data.

% \subsection{Datasets}
% \label{subsec:datasets}
% We experiment with two datasets: 1) translations of WikiQA~\cite{yang2015wikiqa}, an English AS2 dataset, and 2) \tydiastwo introduced in \S~\ref{subsec:tydias2}.

\subsection{AS2 Models}
\label{subsec:as2_models}
This section describes Transformer~\cite{vaswani2017attention} models we use as AS2 models.
Table~\ref{table:pretrained_models} shows the full list of Hugging Face Transformers~\cite{wolf2020transformers} pretrained language models used in this study.

Table~\ref{table:pretrained_models} summarizes the pretrained language models used in this study.
We note that the teacher models in Table~\ref{table:pretrained_models} are fine-tuned on the original English corpus in the target datasets, thus there are two ELECTRA-Large models separately fine-tuned to be teachers for \xtrwikiqa and \tydiastwo.

\subsubsection{English Teacher Models}
To ensure non-specificity to a particular teacher, we experiment with two English AS2 models as teachers in CLKD for \xtrwikiqa: RoBERTa-Large~\cite{liu2019roberta} and ELECTRA-Large~\cite{clark2019electra} are the teacher models trained by~\cite{matsubara2022ensemble} for WikiQA using TANDA, a state-of-the-art AS2 model training method~\cite{garg2020tanda}.
For \tydiastwo, we use the ELECTRA-Large model fine-tuned by TANDA on the \tydiastwo English dataset instead of WikiQA as the teacher.

\subsubsection{Student Models}
We experiment with both monolingual and multilingual pretrained language models (PLMs) as students. Additionally, while we experiment with monolingual training for both the two types of student PLMs, for multilingual students, we also experiment with multilingual training using data for all the languages in the corresponding dataset. 

\paragraph{Monolingual Student Models}
For experiments with \xtrwikiqa, we use ELECTRA-Base~\cite{clark2019electra} pretrained on Hindi corpus and BERT-Base~\cite{devlin2019bert} pretrained on Arabic, German, Italian, Japanese, Dutch, and Portuguese, respectively. We did not find working monolingual PLMs for Spanish and French.
For \tydiastwo, we use ELECTRA-Base~\cite{clark2019electra} pretrained on Bengali corpus, mBERT~\cite{devlin2019bert} finetuned on Swahili corpus, and BERT-Base pretrained on Finnish, Indonesian, Japanese, Korean, and Russian respectively.
Table~\ref{table:pretrained_models} includes the pretrained monolingual models used in this study.

\paragraph{Multilingual Student Models}
As pretrained multilingual student models, we use mBERT~\cite{devlin2019bert}, XLM-RoBERTa-Base (XLM-R-Base), and XLM-RoBERTa-Large (XLM-R-Large)~\cite{conneau2020unsupervised}.

\subsection{Training Languages}
\label{subsec:training-langs}
In addition to pretrained monolingual and multilingual student models, we also experiment with mono- and multilingual training.
For monolingual training, we train the model using a single language's training data. We refer to this setting as \single. For the multilingual setting, which is only possible for multilingual models, we use data for all the languages in a particular dataset, which we refer to as \all.

\subsection{Methods}
\label{subsec:training-methods}
For each dataset, student model, and training languages (\single or \all), we compare two approaches: direct finetuning using gold labels and CLKD using a teacher's soft labels. We refer to these as \finetune and \clkd, respectively. 
In particular, we use \clkde and \clkdr to denote \clkd with ELECTRA-Large and RoBERTa-Large as English teachers, respectively.
Finally, for experiments using original language data (\emph{i.e.}, with \tydiastwo), we additionally compare with the MT-English AS2 pipeline, which involves directly feeding the English translations of the test instances to the English Teacher. This is considered as a strong baseline in~\cite{asai-etal-2021-xor} and~\cite{li2022learning}.

Note that a potential baseline could be to translate all the English data that the teacher was trained on to the target language. However, this is not a feasible approach as (i) the data may not be available, and (ii) even if it were, it would be prohibitively expensive to translate and retrain for every language. Moreover, it will still suffer from the shortcomings of training on translationese such as artifacts from MT and cultural bias as described in \S~\ref{subsec:clkd}.

\subsection{Training Details}
\label{subsec:training-details}
For every model and training configuration, we run three training sessions with different random seeds and present average results.
Our implementation is based on PyTorch~\cite{paszke2019pytorch} and Hugging Face Transformers~\cite{wolf2020transformers} with Python 3.7.

Unless specified otherwise, we use the same hyperparameters for both the supervised baselines and CLKD.
To train AS2 models, we use AdamW~\cite{loshchilov2018decoupled} with an initial learning rate of $10^{-6}$ and the linear scheduler with a warm-up for the first 2.5\% of the training iterations. The number of training iterations (model updates) is set to 20,000 and 40,000 for \xtrwikiqa and \tydiastwo, respectively. For better training convergence with multilingual training, we increase the number of training iterations to 150,000. We use only 1 GPU to train each of the AS2 models.

In this study, we select $\tau \in \{1, 3, 5, 7\}$ based on the results for the development split\footnote{Tables~\ref{table:wikiqa-best-dev-temp}-~\ref{table:tydias2-original-best-dev-temp} in Appendix~\ref{sec:clkd-temp} show the selected temperatures.} and report the averaged test results of the selected AS2 model individually trained with three different random seeds.
To run the extensive amount of experiments, we use Amazon EC2\footnote{\url{https://aws.amazon.com/ec2/}} instances of \texttt{p2.8xlarge}, \texttt{p3.8xlarge}, and \texttt{p3dn.24xlarge}.

%% file: sections/6-results.tex
\section{Evaluations}
\label{sec:evaluations}
We now describe the results of our experiments.
In \S~\ref{subsec:results-translationese}, we have a problem setting where we assume that no target language data is available for training and we use translations of the English data instead.
In \S~\ref{subsec:results-original}, we use the \tydiastwo dataset to experiment with the setting where some original target language unlabeled data is available.

\subsection{Translationese}
\label{subsec:results-translationese}
Tables~\ref{table:wikiqa} and~\ref{table:tydias2-translationese} show the results for all the experiments with the \xtrwikiqa and \xtrtydiastwo translationese datasets, respectively.
Note that the experiments with the \xtrtydiastwo dataset use the test split of \tydiastwo for the evaluation.
It is clear that the performance improves with increasing student model size and going from monolingual training to multilingual training in all languages, even though the training datasets for the diverse languages are translationese from the English corpus.
Nevertheless, CLKD consistently outperforms supervised finetuning with gold labels for all the target languages in both the datasets, and we confirm that \clkd significantly improves \finetune on nearly all the considered configurations of teacher, student, and both monolingual and multilingual training.
This is true even for the \xtrtydiastwo dataset, which is nearly eight times larger than \xtrwikiqa, making it even more challenging to reach the performance of the supervised baseline (\finetune) with an unsupervised method. 

Finally, the performance improvement is greater for smaller models and for monolingual training. This is expected as the teachers have also been trained on the source English dataset, and their performances can be seen as upper bounds for the student performances. The results demonstrate the benefits of soft labels from a strong English AS2 teacher when training AS2 models with no original language data. 

\subsection{Original Language Data}
\label{subsec:results-original}
Table~\ref{table:tydias2-original} shows results of experiments with the original language datasets in \tydiastwo.
We observe similar trends as with translationese; the performance improves with bigger models and multilingual training, and CLKD clearly rivals and regularly outperforms the supervised finetuning with gold-labeled target language data.
These results are especially surprising for a method that does not require any manual annotation as \tydiastwo has diverse questions and documents in native languages and orders of magnitude more data than \xtrwikiqa does. 

Unlike translationese, however, the English teacher's performance is no longer an upper bound as the student models are trained on original language data. In fact, the XLM-R-Large model with both the supervised finetuning and CLKD using data from all the languages (\all) consistently outperforms the \textsc{MT+Teacher} pipeline for all the considered target languages.
The improved results in Table~\ref{table:tydias2-original} v/s Table~\ref{table:tydias2-translationese} also confirm the importance of training on original target language data as opposed to translationese.

Since the cost of manual annotation makes supervised finetuning infeasible for AS2 tasks, these results demonstrate the advantages of our proposed approach in being able to leverage strong English AS2 models and original target language data. While supervision from a strong English teacher precludes the need for costly manual annotation, when used with original target language data, the teacher absorbs the errors and artifacts introduced by machine translation allowing the student to be trained directly on native text.
Moreover, the soft-label supervision from the teacher seems even more useful than gold labels for mono-lingual training and/or smaller student models.

%% file: sections/7-conclusion.tex
\section{Conclusion}
\label{sec:conclusion}

In this work, we proposed cross-lingual knowledge distillation (CLKD) to leverage strong English AS2 models to train accurate models for low-resource languages without the need for costly manual annotation.
Furthermore, we introduced 1) \xtrwikiqa, a machine-translated WikiQA dataset in 9 additional languages and 2) \tydiastwo, a new multilingual AS2 benchmark spanning 8 languages.

We conducted comprehensive experiments involving various teachers, students, and training settings, to discuss the potential of CLKD.
Our results demonstrate the benefits of using soft supervision from a strong English teacher to train a student model for low-resource languages, suggesting the importance of original target language data compared to translationese potentially due to cultural biases and noise introduced by machine translation.

Despite requiring no manual annotations, CLKD leverages both strong English teachers and original target language data and outperforms or rivals strong baselines such as supervised finetuning with the same amount of data and direct usage of a strong teacher model on English translations.

The results also suggest that CLKD has a potential to greatly reduce the cost of training strong AS2 models for languages lacking labeled training data. To engage studies on AS2 for such languages, we publish \xtrwikiqa$\textsuperscript{\ref{fn:xtrwikiqa-url}}$ and \tydiastwo.$\textsuperscript{\ref{fn:tydias2-url}}$

%% file: sections/8-required.tex
\section*{Limitations}
The proposed CLKD is technically applicable to other NLP tasks, but we discuss the effectiveness of the approach for question answering systems, specifically for answer sentence selection (AS2) tasks.
In this study, we put our focus on AS2 tasks as the research community has not well discussed or proposed multilingual AS2 tasks/datasets.
We also find that using only English teacher models is another major limitation of this study.
However, choices of teacher models in the proposed CLKD are not limited to English models.
It would be interesting to discuss the generalizability of the proposed CLKD beyond AS2 tasks, but we note that such discussions would need a much more space to conduct as comprehensive experiments for different NLP tasks as we did for AS2 tasks.

%% file: tables/wikiqa-best_dev_temp.tex
\begin{table*}[t]
    \centering
    \begin{tabular}{lllrrrrrrrrr}
        \toprule
        \multicolumn{1}{c}{\textbf{Language}} & \multicolumn{1}{c}{\textbf{Student LM}} & \multicolumn{1}{c}{\textbf{Method}} & \multicolumn{1}{c}{\textbf{ar}} & \multicolumn{1}{c}{\textbf{de}} & \multicolumn{1}{c}{\textbf{es}} & \multicolumn{1}{c}{\textbf{fr}} & \multicolumn{1}{c}{\textbf{hi}} & \multicolumn{1}{c}{\textbf{it}} & \multicolumn{1}{c}{\textbf{ja}} & \multicolumn{1}{c}{\textbf{nl}} & \multicolumn{1}{c}{\textbf{pt}} \\
        \midrule
        \single & Monolingual & \clkde & 7 & 5 & N/A & N/A & 1 & 5 & 5 & 7 & 5 \\
        ~ & ~ & \clkdr & 7 & 7 & N/A & N/A & 1 & 3 & 5 & 3 & 3 \\
        \cline{2-12}
        ~ & mBERT & \clkde & 5 & 3 & 7 & 5 & 5 & 5 & 7 & 3 & 7 \\
        ~ & ~ & \clkdr & 3 & 7 & 7 & 3 & 5 & 5 & 3 & 7 & 1 \\
        \cline{2-12}
        ~ & XLM-R-Base & \clkde & 7 & 7 & 5 & 5 & 5 & 5 & 7 & 5 & 5 \\
        ~ & ~ & \clkdr & 7 & 7 & 5 & 3 & 5 & 5 & 5 & 1 & 3 \\
        \cline{2-12}
        ~ & XLM-R-Large & \clkde & 3 & 5 & 3 & 7 & 3 & 7 & 5 & 3 & 5 \\
        ~ & ~ & \clkdr & 7 & 7 & 5 & 5 & 3 & 7 & 3 & 3 & 7 \\
        \hline
        \all & mBERT & \clkde & 3 & 5 & 3 & 3 & 3 & 3 & 3 & 5 & 5 \\
        ~ & ~ & \clkdr & 7 & 5 & 3 & 3 & 3 & 3 & 5 & 5 & 7 \\
        \cline{2-12}
        ~ & XLM-R-Base & \clkde & 3 & 5 & 7 & 1 & 5 & 5 & 3 & 3 & 5 \\
        ~ & ~ & \clkdr & 7 & 5 & 5 & 5 & 3 & 3 & 3 & 5 & 7 \\
        \cline{2-12}
        ~ & XLM-R-Large & \clkde & 7 & 7 & 7 & 3 & 1 & 5 & 5 & 5 & 7 \\ 
        ~ & ~ & \clkdr & 3 & 7 & 3 & 1 & 5 & 5 & 7 & 7 & 5 \\
        \bottomrule
    \end{tabular}
    \caption{\textbf{\xtrwikiqa}: Best temperature $\tau$ for \clkd we found in search space (see \S~\ref{subsec:training-details}) with respect to dev split and used to report test results in Table~\ref{table:wikiqa}.}
    \label{table:wikiqa-best-dev-temp}
\end{table*}

%% file: appendices/dataset-validation.tex
\section{Dataset Validation}
\label{sec:dataset-validation}
Since sentence tokenization and identifying answer sentences can introduce errors, we conducted a manual validation of the \tydiastwo datasets.
For each language, we randomly selected 50 instances and verified the accuracy of the answer sentences through manual inspection.
Our findings revealed that the answer sentences were accurate in 98\% of the cases.

%% file: appendices/best-hyperparam.tex
\section{Temperatures for CLKD}
\label{sec:clkd-temp}

Tables~\ref{table:wikiqa-best-dev-temp} -~\ref{table:tydias2-original-best-dev-temp} present the best hyperparameter value of temperature $\tau$ in CLKD for each configuration for \xtrwikiqa, \xtrtydiastwo, and \tydiastwo datsaets.
Following~\cite{matsubara2022ensemble}, we select the best temperature value in terms of mean average precision for the development split.
Those hyperparameter values are used to obtain student models presented in Tables~\ref{table:wikiqa} -~\ref{table:tydias2-original}.

%% file: tables/tydias2-translationese-best_dev_temp.tex
\begin{table}[t]
    \centering
    \small
    \bgroup
    \setlength{\tabcolsep}{0.2em}
    \def\arraystretch{1.2}
    \begin{tabular}{lllrrrrrrr}
        \toprule
        \multicolumn{1}{c}{\textbf{Language}} & \multicolumn{1}{c}{\textbf{Student LM}} & \multicolumn{1}{c}{\textbf{Method}} & \multicolumn{1}{c}{\textbf{bn}} & \multicolumn{1}{c}{\textbf{fi}} & \multicolumn{1}{c}{\textbf{id}} & \multicolumn{1}{c}{\textbf{ja}} & \multicolumn{1}{c}{\textbf{ko}} & \multicolumn{1}{c}{\textbf{ru}} & \multicolumn{1}{c}{\textbf{sw}} \\
        \midrule
        \single & Monolingual & \clkde & 1 & 3 & 3 & 3 & 3 & 3 & 3 \\
        \cline{2-10}
        ~ & mBERT & \clkde & 1 & 3 & 3 & 5 & 3 & 3 & 3 \\
        \cline{2-10}
        ~ & XLM-R-Base & \clkde & 3 & 3 & 5 & 3 & 3 & 3 & 1 \\
        \cline{2-10}
        ~ & XLM-R-Large & \clkde & 3 & 5 & 5 & 3 & 1 & 5 & 3 \\
        \hline
        \all & mBERT &  \clkde & 3 & 3 & 3 & 3 & 3 & 3 & 1 \\
        \cline{2-10}
        ~ & XLM-R-Base & \clkde & 3 & 3 & 3 & 3 & 5 & 3 & 1 \\
        \cline{2-10}
        ~ & XLM-R-Large & \clkde & 5 & 7 & 7 & 7 & 1 & 5 & 7 \\
        \bottomrule
    \end{tabular}
    \egroup
    \caption{\textbf{\xtrtydiastwo (\uline{translationese})}: Best temperature $\tau$ for \clkd we found in search space (see \S~\ref{subsec:training-details}) with respect to dev split and used to report test results in Table~\ref{table:tydias2-translationese}.}
    \label{table:tydias2-translationese-best-dev-temp}
\end{table}

%% file: tables/tydias2-original-best_dev_temp.tex
\begin{table}[t]
    \centering
    \small
    \bgroup
    \setlength{\tabcolsep}{0.2em}
    \def\arraystretch{1.2}
    \begin{tabular}{lllrrrrrrr}
        \toprule
        \multicolumn{1}{c}{\textbf{Language}} & \multicolumn{1}{c}{\textbf{Student LM}} & \multicolumn{1}{c}{\textbf{Method}} & \multicolumn{1}{c}{\textbf{bn}} & \multicolumn{1}{c}{\textbf{fi}} & \multicolumn{1}{c}{\textbf{id}} & \multicolumn{1}{c}{\textbf{ja}} & \multicolumn{1}{c}{\textbf{ko}} & \multicolumn{1}{c}{\textbf{ru}} & \multicolumn{1}{c}{\textbf{sw}} \\
        \midrule
        \single & Monolingual & \clkde & 1 & 3 & 1 & 3 & 5 & 3 & 3 \\
        \cline{2-10}
        ~ & mBERT & \clkde & 1 & 3 & 3 & 3 & 1 & 3 & 3 \\
        \cline{2-10}
        ~ & XLM-R-Base & \clkde & 3 & 3 & 3 & 3 & 3 & 3 & 3 \\
        \cline{2-10}
        ~ & XLM-R-Large & \clkde & 1 & 1 & 3 & 1 & 3 & 3 & 3 \\
        \hline
        \all & mBERT &  \clkde & 1 & 3 & 1 & 3 & 1 & 3 & 3 \\
        \cline{2-10}
        ~ & XLM-R-Base & \clkde & 3 & 1 & 3 & 3 & 1 & 3 & 1 \\
        \cline{2-10}
        ~ & XLM-R-Large & \clkde & 1 & 3 & 3 & 5 & 1 & 3 & 1 \\
        \bottomrule
    \end{tabular}
    \egroup
    \caption{\textbf{\tydiastwo (\uline{original})}: Best temperature $\tau$ for \clkd we found in search space (see \S~\ref{subsec:training-details}) with respect to dev split and used to report test results in Table~\ref{table:tydias2-original}.}
    \label{table:tydias2-original-best-dev-temp}
\end{table}